\documentclass[12pt,a4paper]{article}
\usepackage[utf8]{inputenc}

\usepackage{verbatim}
\usepackage{hyperref}
\usepackage{algorithm,algcompatible,algpseudocode}
\usepackage{amsmath,amssymb,amsthm,amsfonts}
\usepackage[titletoc]{appendix}
\usepackage{tikz}
\usepackage{txfonts}
\usepackage{enumerate}
\usepackage{hyperref}
\usepackage{graphicx}
\usepackage{tcolorbox}
\usepackage{breakcites}
\usepackage[nottoc]{tocbibind}

\hypersetup{bookmarksnumbered}
\graphicspath{ {images/} }

\newtheorem{hypothesis}{Hypothesis}

\newtheorem{definition}{Definition}

\title{\textbf{Multi-Agent Reinforcement Learning: A Report on Challenges and Approaches}}
\author{Sanyam Kapoor\\ \texttt{sanyam@nyu.edu}}
\date{\today}

\begin{document}
\maketitle

\begin{abstract}
Reinforcement Learning (RL) is a learning paradigm concerned with learning to control a system 
so as to maximize an objective over the long term. This approach to learning has received
immense interest in recent times and success manifests itself in the form of human-level 
performance on games like \textit{Go}. While RL is emerging as a practical component in
real-life systems, most successes have been in Single Agent domains. This report will
instead specifically focus on challenges that are unique to Multi-Agent Systems interacting
in mixed cooperative and competitive environments. The report concludes with advances in the
paradigm of training Multi-Agent Systems called \textit{Decentralized Actor, Centralized Critic}, 
based on an extension of MDPs called \textit{Decentralized Partially Observable MDP}s, which 
has seen a renewed interest lately.
\end{abstract}

\cleardoublepage
\tableofcontents
\listoffigures

\cleardoublepage
\section{Introduction}

\subsection{The Reinforcement Learning Problem}

\begin{figure}[ht]
\centering
\includegraphics[width=3in]{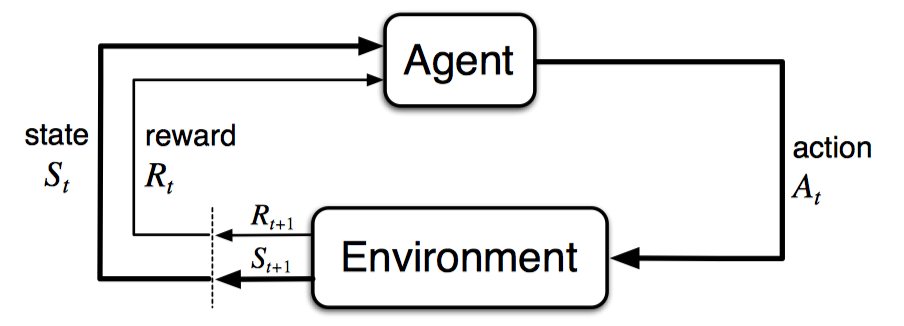}
\caption{The agent-environment feedback loop \cite{Sutton:1998:IRL:551283}} \label{fig:agent_environment_feedback}
\end{figure}

Reinforcement Learning (RL) refers to both the learning problem and sub-field of machine 
learning. The learning problem is to control a system so as to maximize a numerical value which 
represents a long-term objective. The learner is known as the \textit{agent} and everything 
outside the agent is known as the \textit{environment}. The \textit{agent} selects \textit{actions}
and the \textit{environment} responds by presenting a \textit{reward} and a new \textit{state}. A
canonical view of this feedback loop is shown in Figure \ref{fig:agent_environment_feedback}.

Reinforcement Learning, in a sense, is the most general formulation of the learning problem. 
Unlike Supervised Learning, the feedback is partial and in many cases the rewards are delayed.
It also differs from Unsupervised Learning because the aim is not to find hidden structure in
unlabeled data but to solely maximize the reward signal. This importance of the reward signal
is embodied by an informal idea known as the \textit{reward hypothesis} \cite{Sutton:1998:IRL:551283}.

\begin{hypothesis}[\textbf{The Reward Hypothesis}] \label{hypothesis:reward}
That all of what we mean by goals and purposes can be well thought of as the maximization of 
the expected value of the cumulative sum of a received scalar signal (called reward).
\end{hypothesis}

\textit{The Reward Hypothesis} is a matter of ongoing discussion and has not yet received a
convincing consensus{\footnote{Some perspectives can be read at \url{http://
incompleteideas.net/rlai.cs.ualberta.ca/RLAI/rewardhypothesis.html}}}. An RL researcher is 
conventionally expected to come up with a good reward function and subsequently provide a 
robust RL algorithm to generalize to unseen trajectories of the feedback loop seen in Figure 
\ref{fig:agent_environment_feedback}. The process of designing rewards for the problem is 
known as \textit{reward shaping}. It has had apparent criticism because deciding the right
reward is a crucial and delicate matter. Exhaustively modeling dependencies in the environment
can become messy very quickly. This limitation has kept the reward functions in the literature to
be relatively simple and easy to interpret. For instance, in a zero-sum game the rewards are
sparse ($+1$ for a win and $-1$ for a loss). In other cases, where time is critical to achieve the
objective, a reward of $-1$ is given until the agent reaches the objective.

In the following discussion, this hypothesis is assumed to be true. This assumption is motivated
by the promise shown in recent works which use fairly simple and interpretable reward
structures. The reader is however invited to read a more formal discussion around \textit{reward
shaping} in \cite{Ng:1999:PIU:645528.657613}.

\subsection{A Note on Evolutionary Computation}

It is worth acknowledging \textit{Evolutionary Computation} (EC) as an alternative formulation
of the problem of learning autonomous agents. This is a family of techniques in which abstract
\textit{Darwinian} models of evolution are applied to refine populations of candidate solutions to
a given problem. The refinement happens when the fitness function is used to modify the
current population with a breed of new individuals via \textit{Genetic Algorithms} (GA) and
\textit{Evolution Strategies} (ES). Interested readers are invited to read \cite{DeJong:2002:EC:1137808}.

EC is different from RL in that there is no explicit interaction between the environment and the
agent (or an individual in EC). The interaction, if any, happens implicitly via the fitness function
leading to selection, mutation or breeding of individuals. As a result, these methods ignore
much of the structure that is available in the Reinforcement Learning setting - they don't
observe the trajectory of states \& actions that an agent goes through to achieve the objective.
The set of trajectories can provide a rich signal to be exploited and potentially generalize 
unseen environments better.

Subsequently, a detailed discussion on EC, the comparison of EC and RL algorithms and the 
idea of hybrid EC-RL based approaches are beyond the scope of this work. Recent work by
\cite{2017arXiv170303864S} discovers that \textit{Evolution Strategies} represent simple hill-
climbing in a high-dimensional space based only on finite differences and might be of interest
to relevant audience.

\subsection{Motivating the Multi-Agent Setting}

The grand vision of Artifical Intelligence since its inception has been to build autonomous
agents that can interact with the environment and amongst each other. RL formulation of the
learning problem for single agents comes the closest to this vision. Most successes in RL
have been in Single Agent domains where the environment stays largely stationary. It is also
promising to see RL being used in large scale systems such as data center cooling 
\cite{deepmind:dccooling}. Nevertheless, progress in Multi-Agent RL Systems is due.

A number of complex problems in today's society can be modeled as a Multi-Agent Learning
problems. A few examples include Multi-robot control \cite{matignon2012coordinated}, 
analysis of social dilemmas \cite{2017arXiv170203037L}, managing air traffic flow
\cite{Agogino:2012:MAM:2124496.2124503} and energy distribution 
\cite{Pipattanasomporn:4840087}. Traditional RL algorithms are poorly suited for such problems
as we will discuss in forthcoming sections.

The \textit{StarCraft II Learning Environment} \cite{vinyals2017starcraft} has emerged to be a
popular testbed for Multi-Agent RL algorithms because it allows for granular control over the
objectives and constraints in the environment map. While being a formidable environment to be
mastered given the enormous complexity in the interactions, my focus recently has been on a
simpler and interpretable environment called Pommerman \cite{pommerman}. This environment
derives from the Bomberman game where four agents compete each other to be the last one
standing (or in a team the last team standing).

The rest of the report organized as follows - Section \ref{section:background} discusses the
key theoretical underpinnings behind Reinforcement Learning. Section \ref{section:rl_control}
expands on the theory to provide a review of approximate approaches to RL problems. Section
\ref{section:challenges} discusses the unique challenges that a Multi-Agent environment faces.
Section \ref{section:approaches} gives a non-exhaustive review of some of the recent 
approaches tackling the multi-agent problem. Section \ref{section:future} presents research
goals and expectations from the future.

\section{Formal Background} \label{section:background}

This section highlights the key definitions and theoretical underpinnings of modern
Reinforcement Learning algorithms.

\subsection{Markov Decision Processes (MDPs)}

\textit{Markov Decision Processes} (MDPs) allow a mathematically idealized formulation of
the reinforcement learning problem and have their origins in dynamical systems. The agent
and environment interact at discrete time steps $t$. At each time step the agent receives a
\textit{state} $S_t$ and decides to take an action $A_t$. The environment responds by giving
a reward $R_{t+1}$ and a new state $S_{t+1}$. Therefore, this repeated process gives rise to
a sequence known as the \textit{trajectory}

\begin{align} \label{eq:sample_trajectory}
S_0,A_0,R_1,S_1,A_1,\dotsc
\end{align}

The \textit{dynamics} of the environment are defined by a probability distribution which defines
the probability that taking an action $A_{t-1}$ in state $S_{t-1}$ will give a reward $R_t$ and
move the agent to state $S_t$. This part of the system is outside the control of agent(s). The
exclusive dependence on just the current state and not the complete history makes this process
Markovian and is crucial to various theoretical properties. We now see a formal definition of
MDPs.

\begin{definition}[\textbf{Markov Decision Process}] \label{def:mdp}
A Markov decision process is a tuple $(\mathcal{S},\mathcal{A}, \mathcal{R}, p)$ such that
\begin{align}
p(s^\prime, r | s, a) = Pr\big\{S_t = s^\prime, R_t = r\text{ }|\text{ }S_{t-1} = s, A_{t-1} = a \big\}
\end{align}
where $S_t \in \mathcal{S}$ (state space), $A_t \in {A}$ (action space), $R_t \in \mathcal{R}$
(reward space) and $p$ defines the dynamics of the process.
\end{definition}

\begin{figure}[ht]
\centering
\includegraphics[width=3in]{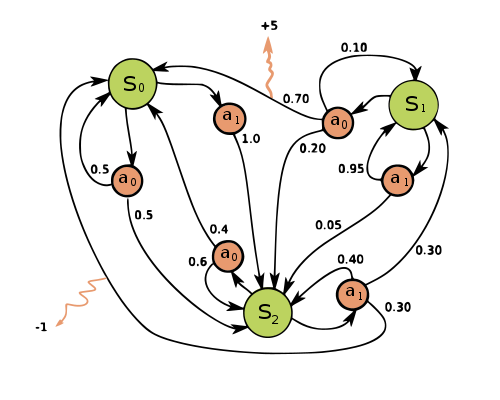}
\caption{Example of a simple MDP with three states (green circles) and two actions (orange 
circles), with two rewards (orange arrows). \textit{by Waldoalvarez distributed under a CC-BY 4.0 
license}} \label{fig:mdp_example}
\end{figure}

Before we go on to define the objective of a Reinforcement Learning agent, we define the notion
of returns. \textit{Returns} are one way of capturing the ``long-term" objective of an RL agent. 

\begin{definition}[\textbf{Discounted Returns}] \label{def:discounted_returns}
Discounted Return is defined as the total sum of rewards following a time step $t$ until the end 
of the sequence of rewards discounted by a factor $\gamma$ at each time step
\begin{align}
G_t &= R_{t+1} + \gamma R_{t+2} + \gamma^2 R_{t+3} + \dotsc \\ \nonumber
G_t &= R_{t+1} + \gamma G_{t+1}
\end{align}
where $R_i \in \mathcal{R}\text{ }\forall i$ and $\gamma \in [0, 1]$.
\end{definition}

The notion of discounted returns is important for it being a good objective for the ``long-term"
because it controls the degree of importance of future rewards for the current time step. An
agent which tries to maximize the objective with $\gamma = 0$ is called ``myopic" and
$\gamma = 1$ gives us the undiscounted returns, being more sensitive to changes in rewards
from the future.

It is important to note that $\gamma < 1$ for \textit{continuing} tasks because the \textit{returns}
must not diverge. We will use \textit{returns} and \textit{discounted returns} interchangeably
from now onwards.

\subsection{Value Functions}

We are now ready to define two imperative value functions which serve as the objective of 
almost all reinforcement learning algorithms.

\begin{definition}[\textbf{Policy}]
A policy is defined as the probability distribution of actions at a given states.
\begin{align}
\pi(A_t = a\text{ }|\text{ }S_t=s)\text{ }\forall S_t \in \mathcal{S}
\end{align}
where $A_t \in \mathcal{A}(s)$ is the state specific action space.
\end{definition}

As with any probability distribution, $\sum_{a} \pi(A_t = a\text{ }|\text{ }S_t=s) = 1$. When the
agent follows a \textit{policy}, it gives rise to a trajectory as seen in Equation 
\ref{eq:sample_trajectory}.

\begin{definition}[\textbf{State Value Function}]
Value function of a state $s$ under policy $\pi$ is defined as the expected return when starting
in state $s$ and following a policy $\pi$ to take actions
\begin{align}
V^\pi(s) = \mathbb{E}_\pi \left[ G_t\text{ }|\text{ }S_t=s \right]\text{ }\forall s \in \mathcal{S}
\end{align}
\end{definition}

\begin{definition}[\textbf{Action Value Function}]
Value function of a state $s$ and action $a$ under policy $\pi$ is defined as the expected return
when starting in state $s$, taking action $a$ and following a policy $\pi$ to take actions further.
\begin{align}
Q^\pi(s,a) = \mathbb{E}_\pi \left[ G_t\text{ }|\text{ }S_t=s, A_t=a \right]\text{ }\forall s \in \mathcal{S}, a \in \mathcal{A}(s)
\end{align}
\end{definition}

\subsection{Bellman Equations}

The \textit{value functions} defined above take a nice recursive form which are defined by
the \textit{Bellman} Equations \cite{Bellman:1957} from Dynamic Programming literature.

\begin{definition} \label{def:bellman_state_value}
The Bellman Expectation Equation for $V^\pi(s)$ is given by
\begin{align}
V^\pi(s) &= \mathbb{E}_\pi \left[ G_t\text{ }|\text{ }S_t=s \right] \\ \nonumber
&= \mathbb{E}_\pi \left[ R_{t+1} + \gamma G_{t+1}\text{ }|\text{ }S_t=s \right] \\ \nonumber
&= \sum_{a} \pi(a|s) \left[ \sum_{s^\prime, r} p(s^\prime,r | s, a) \left[ r + \gamma \mathbb{E}\left[G_{t+1}|S_{t+1}=s^\prime \right] \right] \right] \\ \nonumber
&= \sum_{a} \pi(a|s) \left[ \sum_{s^\prime, r} p(s^\prime,r | s, a) \left[ r + \gamma V^\pi(s^\prime) \right] \right] \\ \nonumber
&= \mathbb{E}_\pi \left[ R_{t+1} + \gamma V^\pi(S_{t+1})\text{ }|\text{ }S_t=s \right]
\end{align}
\end{definition}

The above expectation considers all possibilities of actions by the policy and the induced 
states by those actions defined by the environment dynamics. Similarly, we have

\begin{definition} \label{def:bellman_action_value}
The Bellman Expectation Equation for $Q^\pi(s,a)$ is given by
\begin{align}
Q^\pi(s,a) &= \mathbb{E}_\pi \left[ G_t\text{ }|\text{ }S_t=s, A_t=a \right] \\ \nonumber
&= \mathbb{E}_\pi \left[ R_{t+1} + \gamma Q^\pi(S_{t+1},A_{t+1})\text{ }|\text{ }S_t=s, A_t=a \right]
\end{align}
\end{definition}

It turns out, that for a \textit{finite} MDP, both the Bellman equations above have a unique
solution that can be solved by a system of linear equations defined recursively in definitions
\ref{def:bellman_state_value} and \ref{def:bellman_action_value}. A concise matrix solution is
given in Equation \ref{eq:bellman_solution}.

\begin{align} \label{eq:bellman_solution}
\mathbf{V}^\pi &= \mathcal{R}^\pi + \gamma \mathcal{P}^\pi \mathbf{V}^\pi \\
\implies \mathbf{V}^\pi &= (\mathbf{I} - \gamma \mathcal{P}^\pi)^{-1} \mathcal{R}^\pi
\end{align}

where $\mathbf{V}^\pi$ represents the value vector for each node in a Markov decision 
process under a policy $\pi$ and $\mathcal{P}^\pi$ represents the state transition
matrix. It should be noted that the runtime of this form is prohibitive in practice 
($\mathcal{O}(n^3)$) and we will see practical solutions in \S\ref{section:rl_control}. This also
forms the solution for action value because of the following relation

\begin{align}
V^\pi(s) = \sum_{a} \pi(a|s) Q^\pi(s,a)
\end{align}

Hence, the objective of an RL agent is to maximize these expected value objectives. The next
two optimality equations form the basis of solving the control problem in Reinforcement
Learning as we discuss in the further sections.

\begin{definition} \label{def:bellman_opt_state_value}
The Bellman Optimality Equation for $V^\star(s)$ is given by
\begin{align}
V^\star(s) &= \underset{\pi}{max}\text{ }V^\pi(s) \\ \nonumber
&= \underset{a}{max}\text{ } Q^\star(s,a) \\ \nonumber
&= \underset{a}{max}\text{ } \sum_{s^\prime, r} p(s^\prime,r | s, a) \left[ r + \gamma V^\star(s^\prime) \right]
\end{align}
\end{definition}

\begin{definition} \label{def:bellman_opt_action_value}
The Bellman Optimality Equation for $Q^\star(s,a)$ is given by
\begin{align}
Q^\star(s,a) &= \underset{\pi}{max}\text{ }Q^\pi(s,a) \\ \nonumber
&= \sum_{s^\prime, r} p(s^\prime,r | s, a) \left[ r + \gamma\text{ }\underset{a^\prime}{max}\text{ }Q^\star(s^\prime,a^\prime) \right]
\end{align}
\end{definition}

An extensive discussion on MDPs is presented in \cite{dimitri2017dynamic}.

\section{Reinforcement Learning and Control} \label{section:rl_control}

As seen previously in matrix form, finite MDPs have a unique solution for the \textit{state-value}
and \textit{action-value} functions. However, the runtime is prohibitive and this section discusses
the general frameworks for practical solutions.

The two classic iterative approaches to solve the Dynamic Programming problems are known
as the \textit{Value Iteration} and \textit{Policy Iteration}. These approaches are only possible
when the environment dynamics are known to the agent and in most modern problems of
relevance, that is not the case. Even when the environment dynamics are known, if the
state or action space grows very large, these tabular methods will not be feasible. Hence, in the
interest of brevity and space, this approach has been omitted from discussion. Interested
readers can refer \cite{Sutton:1998:IRL:551283}.

In the absence of environment dynamics or a large state-action space, function approximators
are imperative. Most Reinforcement Learning techniques today can be broadly classified into
either approximation of the table of state and action values, learning a policy distribution for
each state or a mixture of the two. Since, most of these methods require the calculation of 
expectation over the trajectories induced by a policy, Monte Carlo (MC) methods are used to 
approximate the expectation values.

\subsection{Q-Learning} \label{section:q_learning}

Q-Learning \cite{Watkins1992} has been one of the most influential methods in Reinforcement
Learning. The objective here is to learn the \textit{action-value} function $Q^\pi(s,a)$ for policy
$\pi$ by minimizing the expected loss $\mathcal{L}(\theta)$.

\begin{align}
\mathcal{L}(\theta) = \mathbb{E}_\pi\left[ (Q_{\theta}(s,a) - y)^2 \right]
\end{align}

where $y = r + \gamma\text{ }\underset{a^\prime}{max}\text{ }Q_{\theta^\prime}(s^\prime,a^
\prime)$. $y$ represents the Q-Learning target value. Since, the expectation is not directly
computable, it is approximated by sampling a large number of trajectories following
a suitable coverage policy for exploration like $\epsilon$-greedy or Boltzmann exploration 
\cite{2017arXiv170510257C}. 

Q-Learning in its vanilla form tends to be highly unstable especially in the form where Deep
Neural Networks are used for function approximation. Deep Q-Learning \cite{mnih2015human}
was proposed to overcome this problem by introducing the notion of target network whose
parameters are kept constant for a finite number of training steps and is used to generate the
values for $y$. The other technique used to stabilize the performance and overcome the
problem of catastrophic forgetting is to use an Experience Replay Buffer from which transitions
are sampled at random. This also helps break correlation between the sequential transitions
from trajectories generated by the policy $\pi$.

\subsection{Policy Gradient Methods}

Policy Gradient methods differ from Q-Learning in the sense that they explicitly learn a 
stochastic policy distribution $\pi_\theta$ parametrized by $\theta$. One natural choice for
the objective here is to maximize the expected return over the trajectories induced by the policy
$\pi_\theta$. If we denote the reward of a trajectory $\tau$ generated by policy $\pi_\theta(\tau)$
as $r(\tau)$

\begin{align} \label{eq:policy_grad_objective}
J(\theta) &= \mathbb{E}_{\pi_\theta}\left[ r(\tau) \right] = \int \pi_\theta(\tau)r(\tau) d\tau \\
\nabla J(\theta) &= \int \nabla \pi_\theta(\tau)r(\tau) d\tau \\ \nonumber
&= \int \pi_\theta(\tau) \nabla \log \pi_\theta(\tau)r(\tau) d\tau \\ \nonumber
&= \mathbb{E}_{\pi_\theta}\left[ \nabla \log \pi_\theta(\tau)r(\tau) \right] 
\end{align}

Now, the probability of generating the trajectory is 
\begin{align}
\pi_\theta(\tau) = \mathcal{P}(s_0) \prod_{t=1}^T \pi_\theta(a_t|s_t) p(s_{t+1}|s_t,a_t)
\end{align}

where $\mathcal{P}$ refers to the ergodic distribution of the MDP. In simple terms, it is the
probability of the agent being in some initial state. Taking the $\log$ and the gradient reveals
a surprisingly beautiful result

\begin{align}
\log \pi_\theta(\tau) &= \log \mathcal{P}(s_0) + \sum_{i=1}^T \log \pi_\theta(a_t|s_t)  + \log p(s_{t+1}|s_t,a_t) \\ \nonumber
\nabla \log \pi_\theta(\tau) &= \sum_{i=1}^T \nabla \log \pi_\theta(a_t|s_t) \\ \nonumber
\implies \nabla J(\theta) &= \mathbb{E}_{\pi_\theta}\left[ \nabla \left(\sum_{t=1}^T \log \pi_\theta(a_t|s_t) \right) r(\tau) \right]
\end{align}

The gradient comes out to be independent of the environment dynamics and the ergodic
distribution. This means we can now just run Monte-Carlo simulations and approximate the 
gradient to find the best parameters $\theta^\star$ and the computed gradient is an unbiased
estimator of the true gradient.

It should be further observed that the $r(\tau)$ term in the gradient remains effectively
uninfluenced during the gradient operation. Replacing $r(\tau)$ by $G_t$ (discounted
returns from Definition \ref{def:discounted_returns}) gives us the REINFORCE Algorithm 
\cite{williams1992simple}. This replacement is possible because rewards from the past cannot
influence the rewards in the future. 

\subsection{Actor-Critic Methods} \label{section:ac_methods}

The objective used in Policy Gradient Methods leads to very high variance models and part of
the problem is aggravated by the scale of rewards. If observed closely, the policy gradient is
equivalent to a Maximum Likelihood Estimation (MLE). Data overwhelms the prior. Any erratic
trajectories which produce unusual rewards would cause an unexpected change in the resulting
distribution. To mitigate this problem, an idea that helps reduce the variance is to
instead maximize an objective which keeps track of the relative reward difference. This leads
to an algorithm called REINFORCE with a Baseline by introducing a term in the gradient which
does not induce additional bias. Hence, the gradient becomes

\begin{align}
\nabla J(\theta) &= \mathbb{E}_{\pi_\theta}\left[ \nabla \left(\sum_{t=1}^T \log \pi_\theta(a_t|s_t) \right) (G_t - b) \right]
\end{align}

where $b$ is the introduced baseline.

\begin{align}
\mathbb{E}_{\pi_\theta}\left[ \nabla \left(\sum_{t=1}^T \log \pi_\theta(a_t|s_t) \right)b \right] &= \int  \sum_{t=1}^T \pi_\theta(a_t|s_t) \nabla \log \pi_\theta(a_t|s_t) b d\tau \\ \nonumber
&= \int  \nabla \sum_{t=1}^T \pi_\theta(a_t|s_t) b d\tau \\ \nonumber
&= \int \nabla \pi_\theta(\tau) b d\tau \\ \nonumber
&= b \nabla \int \pi_\theta(\tau) d\tau \\ \nonumber
&= b \nabla 1 = 0
\end{align}

The above calculations show that the addition of a baseline keeps the gradient estimate 
unbiased while decreasing the variance. Often, choosing the right baseline is a challenge in
itself and modern methods resolve to using another parametric value function $V^\omega(s)$
as the baseline which is commonly known as the ``critic". A problem with the objective 
dicussed in \ref{eq:policy_grad_objective} is that it might not be fully differentiable and hence 
we introduce a differentiable surrogate in the form of $Q(s,a)$, the action-value function. An
extended treatment of Policy Gradient methods can be found in \cite{Sutton:1998:IRL:551283}.
Recent methods like A3C \cite{pmlr-v48-mniha16} and PPO family of algorithms 
\cite{2017arXiv170706347S} have been proposed to improve the stability of learned policies.
The difference value of the the objective and the baseline is also known as the
\textit{advantage estimate}.

\subsection{Deterministic Policy Gradients} \label{section:dpg}

Instead of representing policies by a parametric probabilistic distribution $\pi_\theta(a|s)$ that
stochastically selects actions $a$, this approach considers determinstic policies of the form
$a = \mu_\theta(s)$. It turns out that the Determinstic Policy Gradient is a limiting case of
Stochastic Policy Gradients when the variance approaches zero \cite{pmlr-v32-silver14}. This
approach has been shown to outperform Stochastic Policy Gradients discussed earlier in
high dimensional action spaces. Following a similar approach as for Policy Gradients, the
gradient expression is

\begin{align}
\nabla_\theta J(\mu_\theta) = \mathbb{E}\left[ \nabla_\theta \mu_\theta(s) \nabla_a 
Q^\mu(s,a) \big|_{a = \mu_\theta(s)}  \right]
\end{align}

This result is surprising as well because this also does not depend on the environment 
dynamics. The policy is decided by solving the following maximization problem.

\begin{align}
\mu^{k+1}(s) = \underset{a}{argmax}\text{ }Q^{\mu^k}(s,a)
\end{align}

However, this is computationally hard to do at each decision step. Instead, multiple gradients 
are averaged from multiple trajectories to reduce variance in the estimate. This work has since
been extended by \cite{2015arXiv150902971L} to be more stable by using the idea of
Experience Replay Buffer and Target Network updates to keep the target network constant for 
few gradient steps.

\section{Challenges in Multi-Agent Environments} \label{section:challenges}

Section \ref{section:background} and \ref{section:rl_control} provide a concise overview of
modern techniques. However, these have primarily been developed for the Single-Agent case.
Before we see extensions of these techniques to a Multi-Agent environment, it is imporant
to understand the various ways in which Multi-Agent Environments fall beyond the direct scope
of the control algorithms discussed above. Here is a non-exhaustive list of challenges.

\subsection{Joint Action Space} \label{section:joint_action_space}

In its most general form, the MDPs can be extended as a framework for Multi-Agent systems
as a Markov Game \cite{Littman:1994:MGF:3091574.3091594}. The state transitions are 
controlled by the current state and one action from each agent. For an environment with $n$ 
agents

\begin{align}
\mathcal{T}&: \mathcal{S} \times \mathcal{A}_{1} \times \mathcal{A}_2 \times \dotsc \times \mathcal{A}_n \to PD(\mathcal{S}) \\
R_i&: \mathcal{S} \times \mathcal{A}_{1} \times \mathcal{A}_2 \times \dotsc \times \mathcal{A}_n \to \mathcal{R} \\
\pi_{\theta_i}&: \mathcal{S} \times \mathcal{A}_{1} \times \mathcal{A}_2 \times \dotsc \times \mathcal{A}_n \to [0,1]
\end{align}

where $PD(\mathcal{S})$ represents the probability distribution over the resultant state space.
As is apparent, all the routines above are now exponentially dependent on the \textit{action
space}. Any decision made by the policy becomes immediately affected by joint actions taken
by all the actors and considering all of them becomes imperative to maximize the agent reward
$R_i$. This is now \textit{computationally hard}. A concerning result by
\cite{2017arXiv170602275L} shows that for a simple setting of binary actions, the probability of
taking a gradient step in the correct direction decreases exponentially with the number of agents. Formally

\begin{align}
Pr\left[ \langle \hat{\nabla} J, \nabla J \rangle > 0 \right] \propto 0.5^N
\end{align}

where the agent's policy is initialzed to an uninformed policy s.t. $\pi(a = 1|s) = 0.5$, $N$ is
the number of agents and $\hat{\nabla} J$ is the gradient estimate from a single sample.

\subsection{Game-Theoretic Effects} \label{section:game_theoretic_effects}

As we've noted earlier, every MDP has at least one optimal policy and of the given optimal
policies at least one is stationary and deterministic. However, for many Markov games, as 
defined in Section \ref{section:joint_action_space}, there is no deterministic optimal policy that 
is \textit{undominated} because it critically depends on the the behavior of the opponent. The
need for stochasticity arises from the agent's uncertainty in its opponent's moves. An 
illustration for the point can be read in Box \ref{box:rps}.

\begin{figure}[ht]
\centering
\includegraphics[width=3in]{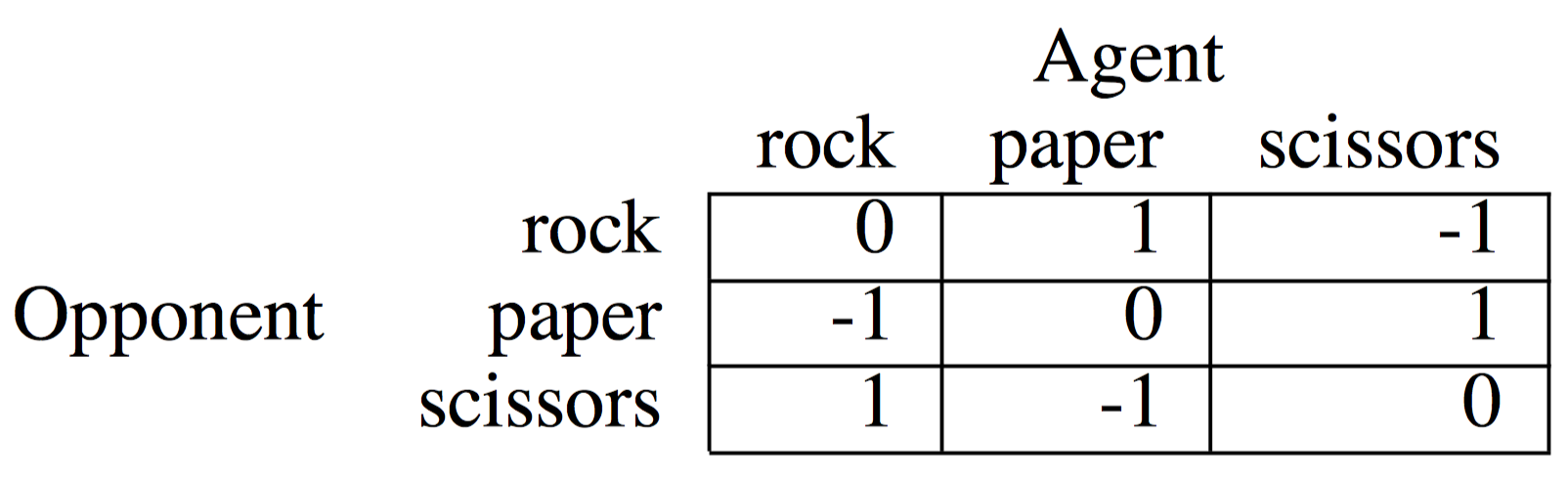}
\caption{The matrix-game for ``Rock, Paper, Scissors" \cite{Littman:1994:MGF:3091574.3091594}} \label{fig:rock_paper_scissors}
\end{figure}

\begin{figure}[ht]
\begin{tcolorbox}[label=box:rps,colback=white]

\textbf{\large{Box \ref{box:rps}: Rock, Paper, Scissors}}

A mathematically convenient way to illustrate stochasticity in Multi-Agent systems is to consider
a zero-sum Markov game called ``Rock, Paper, Scissors" with only one state, also called a 
Matrix game. The objective is to maximize the expected reward. $R_{o,a}$ represents the reward
when agents takes an action $a$ and the opponent takes the action $o$. The game is shown in
Figure \ref{fig:rock_paper_scissors}.

The linear constraints on the problem for expected rewards for a policy $\pi$ and total value
pay-off $V$

\begin{align}
\pi_{\text{paper}} - \pi_{\text{scissors}} \geq V \\
- \pi_{\text{rock}} + \pi_{\text{scissors}} \geq V \\
\pi_{\text{rock}} - \pi_{\text{paper}} \geq V \\
\pi_{\text{rock}} + \pi_{\text{paper}} + \pi_{\text{scissors}} = 1
\end{align}

Linear Programming gives a solution for this as $\pi = (\frac{1}{3}, \frac{1}{3}, \frac{1}{3})$ and
$V = 0$. Hence, the optimal policy for the game is using the random strategy at best! This
can be concisely written in the \textit{maximin} formulation as 

\begin{align}
V = \underset{\pi}{\max}\text{ }\underset{o}{\min} \sum_{a} R_{o,a} \pi_a
\end{align}

\end{tcolorbox}
\end{figure}

While Single-Agent systems have a relatively strong theoretical foundation, a thorough
understanding of the learning problem in \textit{multi-agent} settings is still an open problem.
The transient nature of dynamics in such a setting makes the problem harder to analyze.
\cite{Shoham:2007:MLA:1247754.1248180} strongly cautions to rely not too strongly on
requirements such as convergence to a \textit{Nash Equilibrium} when evaluating learning
algorithms in a multi-agent setting. Instead, \textit{Evolutionary Game Theory} is emerging as
the preferred framework rather than classical game theory and is surveyed in detail by 
\cite{Bloembergen:2015:EDM:2831071.2831085}.

\subsection{Credit Assignment and Lazy Agent Problem}

The \textit{Credit Assignment Problem} concerns with how the success of an overall system can
be attributed to the various contributions of a systems components \cite{Minsky:4066245}. In
a Reinforcement Learning setup, it is already hard to attribute an outcome to a particular action
in history. With the extension to Multi-Agent settings, this problem increases in complexity
mutlifold. It is extremely doubtful that the signal presented by an outcome (e.g. a win or a loss)
contains enough information to make this inference. The naive approach of equally dividing
the outcome reward to each of the agents seldom makes sense. Interested audience can find
a detailed discussion in \cite{Sutton:1984:TCA:911176} and behavorial analysis with different
reward functions in \cite{Balch97learningroles}.

Another phenomena which arises due to partial observability is called the ``Lazy-Agent Problem"
\cite{2017arXiv170605296S} which particularly can occur in Cooperative environments. Learning
can fail when one of the agent becomes inactive because when one agent learns a useful
policy, the second agent can be discouraged from exploration so as to not affect the first 
agent's performance.

\subsection{Non-Markovian Nature of Environments}

The Markov assumption is crucial in the current formulation of RL algorithms. It provides a 
mathematically clean framework to approach the learning problem. However, this
assumption can be easily violated in the simplest of scenarios. Consider the simple case of
an agent learning to find the shortest path from $A$ to $B$. In the unconstrained form, this
problem is \textit{Markovian} however, with a simple addition of the constraint that no 
intermediate states can be revisited makes it non-\textit{Markovian}. 

This problem was  addressed using Recurrent Networks in \cite{schmidhuber1991reinforcement}
to allow the state representation over sequences of state history. Modern treatment of this
problem has followed the same approach to build a hidden representation of state sequence
with Gated Neural Networks or Convolutional Neural Networks.

\section{Decentralized Actor, Centralized Critic} \label{section:approaches}

\begin{figure}[ht]
\centering
\includegraphics[width=3in]{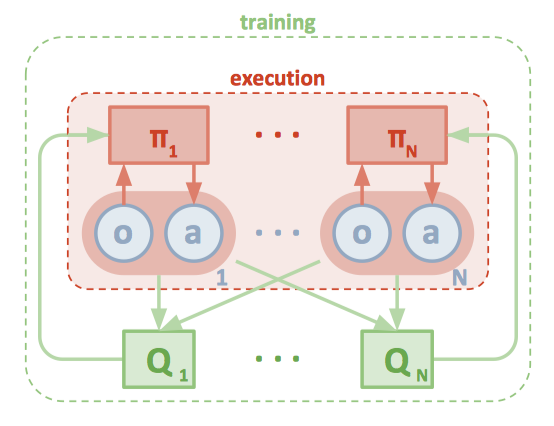}
\caption{Overview of Multi-Agent Decentralized Actor, Centralized Critic approach \cite{2017arXiv170602275L}} \label{fig:dacc}
\end{figure}

In this section, we will take a look at recent approaches to solve the learning  problem in Multi
Agent settings. We will specifically focus on an appealing paradigm of training Reinforcement
Learning Systems for Multiple Agents known as \textit{Decentralized Actor, Centralized Critic}.
The core idea behind this paradigm is summarized in Figure \ref{fig:dacc}.

As seen in Section \ref{section:challenges}, the action space in a Multi-Agent system grows
exponentially with the number of agents. In many cases, learning becomes impossible because
of partial observability and communication constraints. This necessitates the need of a
\textit{decentralized policies} which only depend on local observations of the agents. Such a
formulation naturally attenuates the problem of exponentially growing joint action spaces.

These techniques are augmented in a laboratory setting via a centralized critic which provides
an indirect observation (possibly partial) of the complete global state to each of the actors. This 
helps work around the constraint of inter-agent communications. The primary theoretical
foundation driving work in this region is by an extension of MDPs known as the Decentralized 
Partially Observable Markov Decison Processes (Dec-POMDPs). This framework uses a
generalized notion of states in the form of \textit{observations}. The core idea is that an agent
may only be able to get information over a restricted \textit{horizon}. A formal definition can
be found in \cite{Oliehoek:2016:CID:2967142} with a much detailed discussion.

\cite{kraemer2016multi} present an approach where the agents are allowed to rehearse with
information that will not be available during policy execution and also present weak convergence
guarantees. Following suit, end-to-end Deep architectures have been proposed recently.

\begin{figure}[ht]
\centering
\includegraphics[width=3in]{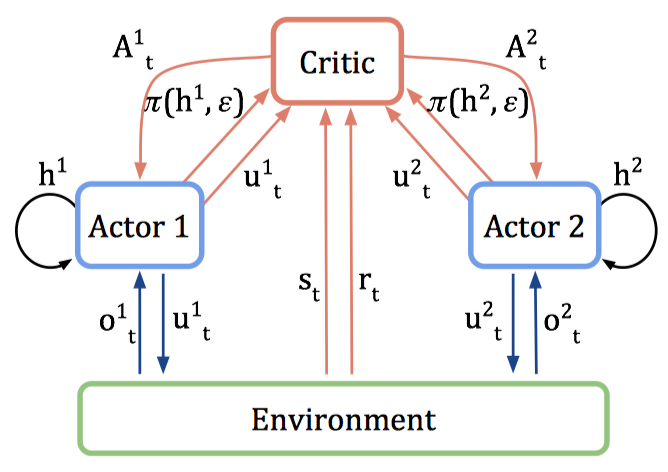}
\caption{Information Flow in COMA \cite{foerster2017counterfactual}} \label{fig:coma}
\end{figure}

One recent approach introduces a clever approach to estimate the \textit{advantage estimate}
by using a counterfactual baseline for policy gradients \cite{foerster2017counterfactual} as

\begin{align}
A^a(s, \mathbf{u}) = Q(s, \mathbf{u}) - \sum_{u^{\prime a}} \pi^a(u^{\prime a} | \tau^a) Q(s,(\mathbf{u}^{-a}, u^{\prime a})
\end{align}

where the advantage estimate is computed for each agent and the baseline marginalizes out
the actions ($u$) of an agent $a$. This allows the centralized critic to reason about the
counterfactuals in which only $a$'s actions change. The interactions between the environment
and the actors are shown in Figure \ref{fig:coma}. $h$ represents the hidden state of the actors
which are existent for the Gated Neural Networks to account for the Non-Markovian nature
of the environments. The training of this network happens via the standard Actor-Critic
approach as seen in Section \ref{section:ac_methods}.

\begin{figure}[ht]
\centering
\includegraphics[width=3in]{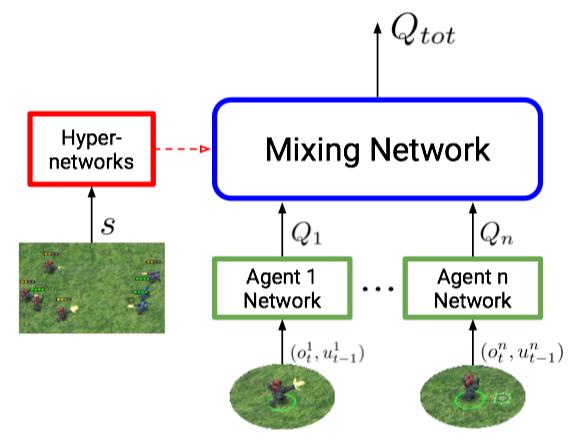}
\caption{Information Flow in QMIX \cite{2018arXiv180311485R}} \label{fig:qmix}
\end{figure}

Another approach applies the same paradigm to Q-Learning by proposing a new objective for
the supervised loss. The key idea is named QMIX \cite{2018arXiv180311485R} which
exploits a linear decomposition of the joint value function across agents by maintaining
monotonicity in the local and global maximum value functions. The information flow is
depicted via Figure \ref{fig:qmix}.

\begin{align}
\frac{\partial Q_{tot}}{\partial Q_a} &\geq 0 \text{ (for all agent's Q-Networks)} \\
\mathcal{L}(\theta) &= \sum_{i} \left[ y_i^{tot} - Q_{tot}(\mathbf{\tau}, \mathbf{u}, s; \theta) \right] \\
y_i^{tot} &= r + \gamma\text{ }\underset{\mathbf{u}^\prime}{\max}\text{ }Q_{tot}(\mathbf{\tau}^\prime, \mathbf{u}^\prime, s^\prime; \theta^{-})
\end{align}

The ``Mixing Network" component of the QMIX architecture in Figure \ref{fig:qmix} is the one
that enforces the monotonicity by taking absolute values of the weights generate by an
auxiliary hyper-network. The end-to-end training of this network happens via the standard
(Deep) Q-Learning framework as seen in Section \ref{section:q_learning}.

By the transient nature of the environment which is critically dependent on the behavior
policies of other interacting agents in the environment, the training becomes highly unstable.
\cite{2017arXiv170602275L} propose to use an ensemble of policies chosen from a pool at
random at the start of each trajectory so that the agents are robust to changes in the 
environment and uses Deterministic Policy Gradients as the choice of training algorithm
as discussed in Section \ref{section:dpg}.

\subsection{Experiments with Pommerman}

A large part of my work currently is about adapting the knowledge above to the novel 
environment of Pommerman \cite{pommerman}. The particular variant of the game I am
interested in solving is the 2v2 variant where two autonomous agents team up against another
team of 2. The paradigm of Multi-Agent training seen above suits exceptionally well to this
environment.

My implementation is inspired by the QMIX architecture seen in Figure \ref{fig:qmix} where the
agents are being trained against a team of rule-based agents. An extensive behavorial analysis of 
the algorithm and tests for robustness are still in progress. It is also to be seen whether these 
agents can be paired up with novel teammates.

\section{Future Work} \label{section:future}

A large set of problems still stay open in both the theoretical and applied aspects of
Reinforcement Learning Systems for Multi-Agent Systems. My immediate next steps
to empirically improve the performance described above will be to incorporate the idea
of ``competitive self-play" \cite{2017arXiv171003748B} and ``kickstarting" agents from a
pool of pre-trained agents \cite{2018arXiv180303835S}. I expect these approaches to perform
better than the current approach of training agents from scratch. I am also optimistic about
\textit{Dec-POMDP}s to be the foundational theoretical framework behind Multi-Agent
Reinforcement Learning and will be exploring this topic further.

\section*{Acknowlegements} \label{section:ack}
\addcontentsline{toc}{section}{\nameref{section:ack}}

I would like to thank Joan Bruna, Roberta Raileanu and Cinjon Resnick for insightful discussions
and directions along the way.

\bibliography{references}

\end{document}